\documentclass[letterpaper, 10 pt, conference]{ieeeconf}  

\IEEEoverridecommandlockouts                              
\usepackage{graphicx}
\usepackage{hyperref}
\usepackage{caption}
\usepackage{subcaption}
\usepackage{float}

\title{\LARGE \bf
A Commute in Data: The \texttt{comma2k19} Dataset
}

\author{Harald Sch{\"a}fer\thanks{Corresponding author's email: harald@comma.ai}, Eder Santana, Andrew Haden, and Riccardo Biasini\\
\texttt{comma.ai}
}

\begin{document}

\maketitle
\thispagestyle{empty}
\pagestyle{empty}

\begin{abstract}
\texttt{comma.ai} presents \texttt{comma2k19}, a dataset of over 33 hours of commute in California's 280 highway. This means 2019 segments, 1 minute long each, on a 20km section of highway driving between California's San Jose and San Francisco. The dataset was collected using comma EONs that have sensors similar to those of any modern smartphone including a road-facing camera, phone GPS, thermometers and a 9-axis IMU. Additionally, the EON captures raw GNSS measurements and all CAN data sent by the car with a comma grey panda. Laika, an open-source GNSS processing library, is also introduced here. Laika produces 40\% more accurate positions than the GNSS module used to collect the raw data. This dataset includes pose (position + orientation) estimates in a global reference frame of the recording camera. These poses were computed with a tightly coupled INS/GNSS/Vision optimizer that relies on data processed by Laika. \texttt{comma2k19} is ideal for development and validation of tightly coupled GNSS algorithms and mapping algorithms that work with commodity sensors. 
\end{abstract}

\section{INTRODUCTION}
``Quality over quantity", or that's what they say anyway, but is this true in the world of data? The reality is that collecting data with high-end sensors is expensive as dedicated hardware is needed and this quickly becomes unfeasible for larger datasets. Affordable sensors on the other hand, are ubiquitous and already continuously logging data on billions of devices around the world. The world is a noisy place, some trends require \textit{big data} to become obvious. To find such trends, algorithms need to be developed to deal with huge amounts of less than perfect data. It is this core idea that motivates \texttt{comma.ai}'s strategy to collect data with scalibility as a priority.

\begin{figure}[t!]
  \centering{}
  \includegraphics[scale=0.35]{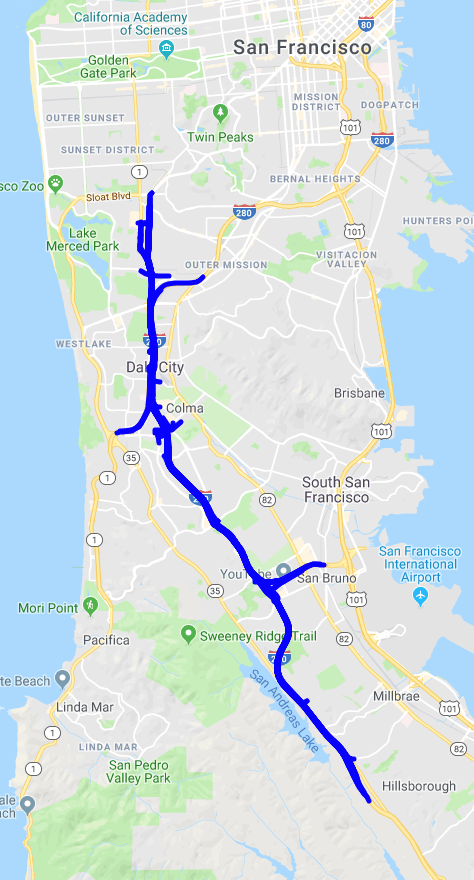}
  \caption{Area driven in the \texttt{comma2k19} dataset.}
  \label{map}
\end{figure}

The dataset released here, \texttt{comma2k19}\footnote{\url{https://github.com/commaai/comma2k19}}, contains data collected by an EON\footnote{\url{https://comma.ai/shop/products/eon-gold-dashcam-devkit/}} and a grey panda\footnote{\url{https://comma.ai/shop/products/panda-obd-ii-dongle/}} during 2019 minutes of driving sampled from a Californian commute (Figure \ref{map}). There are logs of a road-facing camera, a 9-axis IMU, the vehicle's transmitted CAN messages and raw GNSS measurements. This makes this dataset uniquely valuable for the development of mapping algorithms that require dense data and can use raw GNSS data.

Accurate maps are useful for a variety of different applications including surveying, navigation, self-driving cars, etc. Making globally accurate maps requires accurate global pose (position + orientation) of the mapping device/vehicle. Conventionally this is done by fusing global position fixes from a GNSS module with other sensors \cite{schall2009global, rehder2012global, caron2006gps, nemra2010robust}. However, these methods use a pre-computed navigation solution from the GNSS module, i.e. they are \textit{loosely coupled}. A more optimal approach is to directly integrate the raw GNSS measurements into the mapping optimizer/filter, this is called \textit{tight coupling} \cite{angrisano2010gnss, falco2017loose, schreiber2016vehicle}. A tightly coupled GNSS/INS/Vision fusion algorithm is likely the state-of-the-art global pose estimator for a commodity sensor package. The \texttt{comma2k19} dataset is ideal to develop and validate such an algorithm.

We also introduce Laika, an open-source GNSS processing library that was developed and validated using data from the \texttt{comma2k19} dataset. Laika produces significantly more accurate position fixes than reported by the u-blox M8 GNSS module used for raw data collection.

\section{RELATED DATASETS}
There are several driving datasets in the literature such as KITTI \cite{geiger2013vision}, Cityscapes \cite{cordts2016cityscapes}, RoboCar \cite{maddern20171}, ApolloScapes \cite{huang2018apolloscape}, Berkeley Deep Drive \cite{yu2018bdd100k}, including our previous public dataset \cite{santana2016learning}. Most of these datasets focus on high quality sensors such as LIDAR or high level computer vision annotations such as semantic segmentation, object detection and imitation learning.

On the other hand, the dataset presented here focuses on consumer grade  sensors for reproducibility and scalability. Additionally, all the data collected in this dataset is concentrated in a very small area, this high density ensures repeated observations of the same location across a variety of conditions. This combined with the raw GNSS logs makes this dataset more suited for the development of high performance localization and mapping algorithms intended to run on commodity hardware.

\section{Sample choice}
The data was collected with the EON's standard logging infrastructure. This specific highway was chosen because it is representative of the commute of millions of Americans that drive similar urban roads across the country every day. Data was only selected from this small portion of road to ensure that it is sufficiently dense for experiments mapping-related experiments. An interesting challenge of this dataset is that the vision data we collected is quite different from other datasets, in that there are a less \textit{good features to track}\cite{goodfeatures} in the video. This makes it particularly interesting to test vision algorithms that need to work on the common highway driving scenarios.

\section{SENSOR SETUP}
\subsection{Vehicles}
Data was logged on two different setups. A 2016 Honda Civic Touring and a 2017 Toyota RAV4 Platinum. 

\begin{figure}[t!]
  \includegraphics[width=\linewidth]{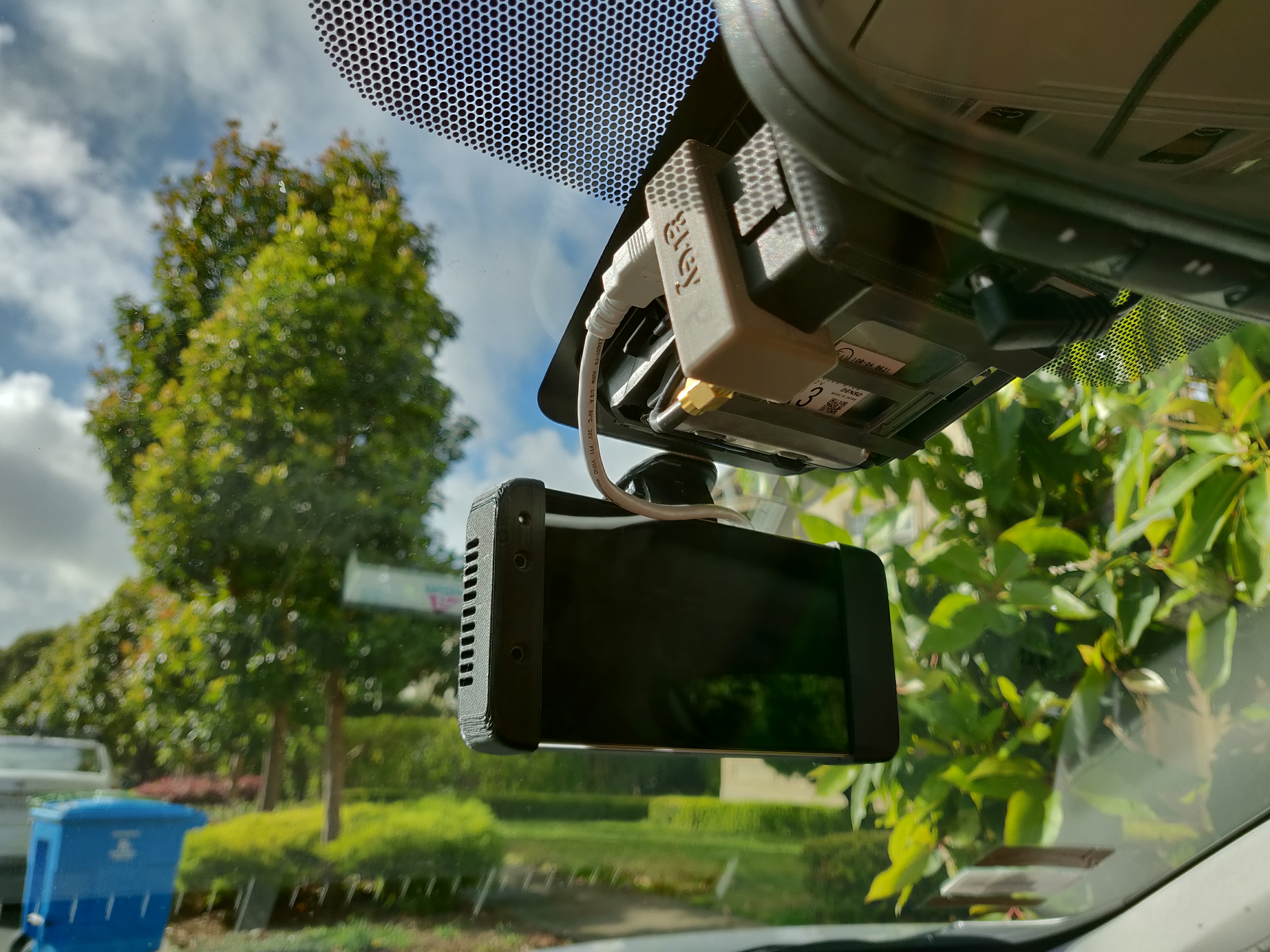}
  \caption{EON and grey panda installed in a vehicle.}
\end{figure}


\subsection{CAN messages}
All the vehicles CAN messages are received and logged. Radar, steering angle and wheel speed readings have been parsed in this dataset.

\subsection{Camera data}
The road-facing camera data was logged with a Sony IMX298\footnote{\url{https://www.sony-semicon.co.jp/products_en/IS/sensor1/products/imx298.html}}  camera sensor. Video is captured at 20Hz and compressed with H.264.

\subsection{Raw GNSS}
The grey panda, contains a u-blox M8 chip\footnote{\url{https://www.u-blox.com/sites/default/files/products/documents/u-blox8-M8_ReceiverDescrProtSpec_\%28UBX-13003221\%29_Public.pdf}} connected to a Tallysman TW4721 antenna. Raw data and u-blox's navigation fix are logged at 10Hz. The raw data includes the doppler shifts, pseudoranges and carrier phases on the L1 channel for GLONASS and GPS.
On the Civic the antenna was mounted inside the car under the windshield, on the RAV4 the antenna was mounted on the roof, resulting in a signal about 15dB stronger.

\subsection{Other Sensors}
Gyro and accelerometer data was collected with a LSM6DS3 at 100Hz and magnetometer data with a AK09911 at 10Hz. The EON also has an integrated WGR7640 GNSS receiver that also logs raw GNSS measurements in the same format as the u-blox module and logs at 1Hz. However, at least partly due to the bad antenna, the quality of the WGR7640 data is much lower.

\begin{figure*}[!htb]
  \centering{}
  \includegraphics[scale=0.55]{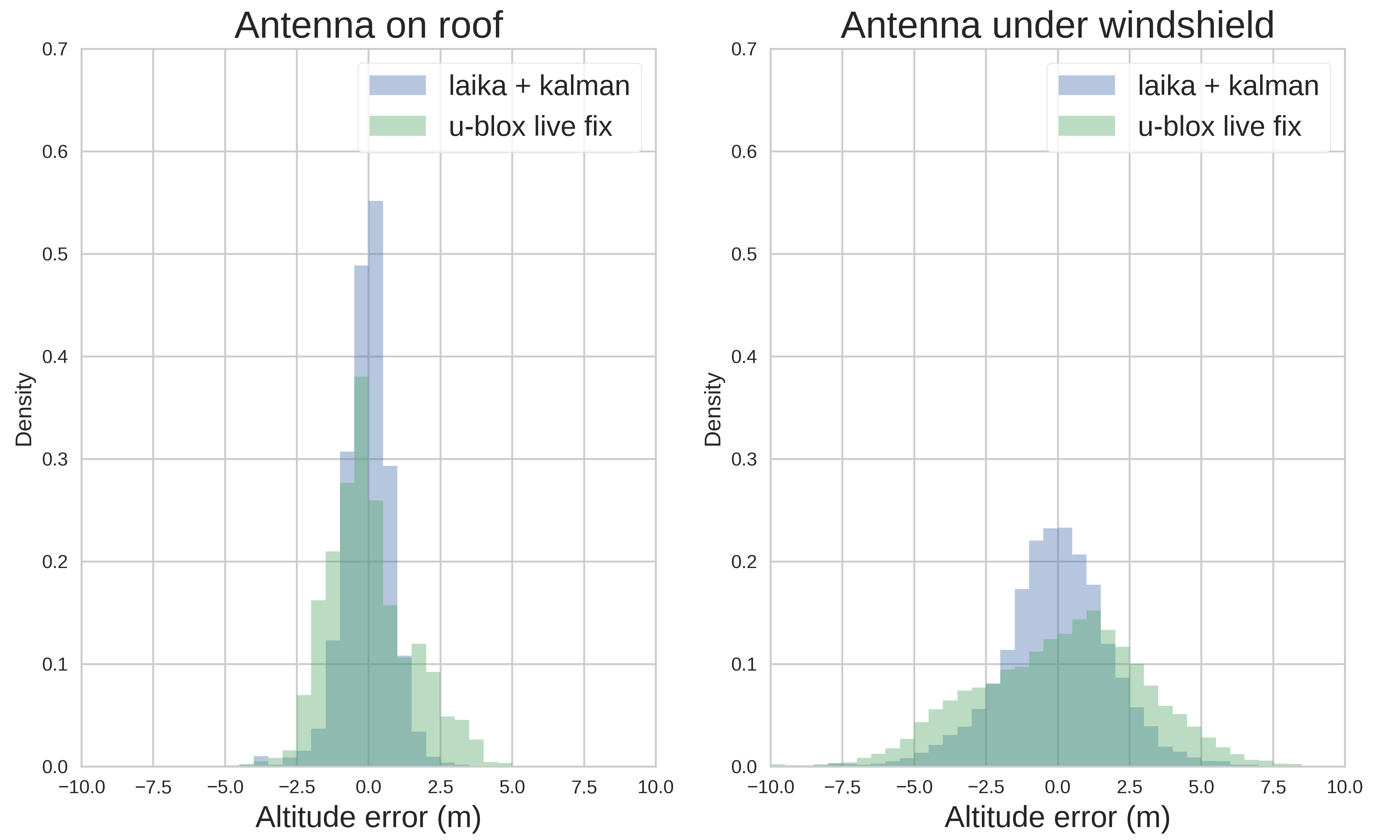}
  \captionsetup{width=.8\linewidth}
  \caption{Altitude error distributions for Laika and Live u-blox baseline algorithm for the two scenarios with antenna on the roof (left) and inside the car (right).}
  \label{laika_accuracy}
\end{figure*}

\section{LAIKA}\label{LAIKA}
Laika\footnote{\url{https://github.com/commaai/laika}} is an open source GNSS processing library developed with \texttt{comma2k19}. Laika is similar to projects like \cite{takasu2011rtklib} and \cite{harris2007gpstk}, with a strong focus on simplicity, readability and straight-forward integration with other optimizers. Laika can be used to compute location fixes from the raw GNSS data that can be significantly more accurate than the live fix computed by GNSS module used for data collection. 

To compute the fixes, raw measurements from the dataset are processed with Laika and then fed into a Kalman filter or an other optimizer that estimates positions. To prove the efficacy of Laika we used a simple Kalman filter that only takes GNSS data as input.  A lack of ground truth can make it difficult to judge GNSS algorithms, since the true position of the receiver is never known. However, assuming the height of the road is constant within a small area, we can estimate the altitude accuracy of a position fix by checking the variation of estimated road height over small sections (5m x 5m) of road. This requires many passes through the same section of road to be reliable; luckily the high density data from this dataset is more than sufficient. Figure \ref{laika_accuracy} shows the altitude error distribution for positions computed with Laika and the positions reported by the u-blox module. Overall the positioning error was reduced by 40\%.

\section{GLOBAL POSES}

\begin{figure}[!htb]
  \centering{}
  \includegraphics[width=\linewidth]{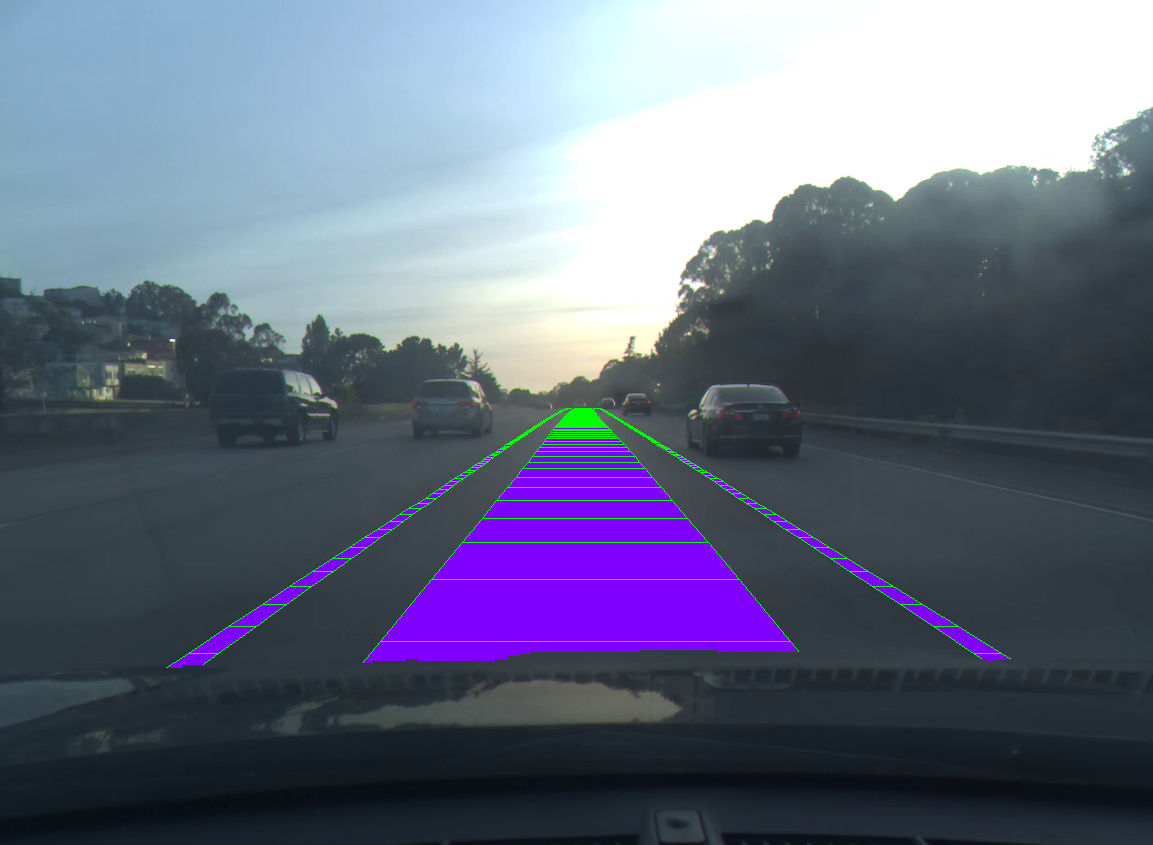}
  \caption{GNSS, INS and vision based 3D estimates of driven path and observed lanes projected onto video.}
  \label{testmesh3d}
\end{figure}

In addition to the raw sensor data, the logs also contain best estimates for global pose (position + orientation) calculated by Mesh3D, \texttt{comma.ai}'s internal post-processing infrastructure that relies on data processed by Laika. They were computed with a tightly coupled GNSS/INS/Vision optimizer, where raw GNSS measurements and ORB \cite{rublee2011orb} features were fed into a Multi-State Constraint Kalman Filter (MSCKF) \cite{mourikis2007multi}, \cite{li2013high}. Figure \ref{testmesh3d} shows a snapshot of the resulting 3D path and lane estimates projected into camera frame.

The global position in the \texttt{comma2k19} is given in ECEF \cite{zhou1999sensor} frame in meters, and the orientation is given as the quaternion that is needed to rotate from ECEF frame into local frame. Where the local frame is defined as $[forward, right, down]$ in accordance with NED (North East Down) \cite{cai2011unmanned} conventions.

\begin{figure*}
  \centering{}
  \includegraphics[scale=0.25]{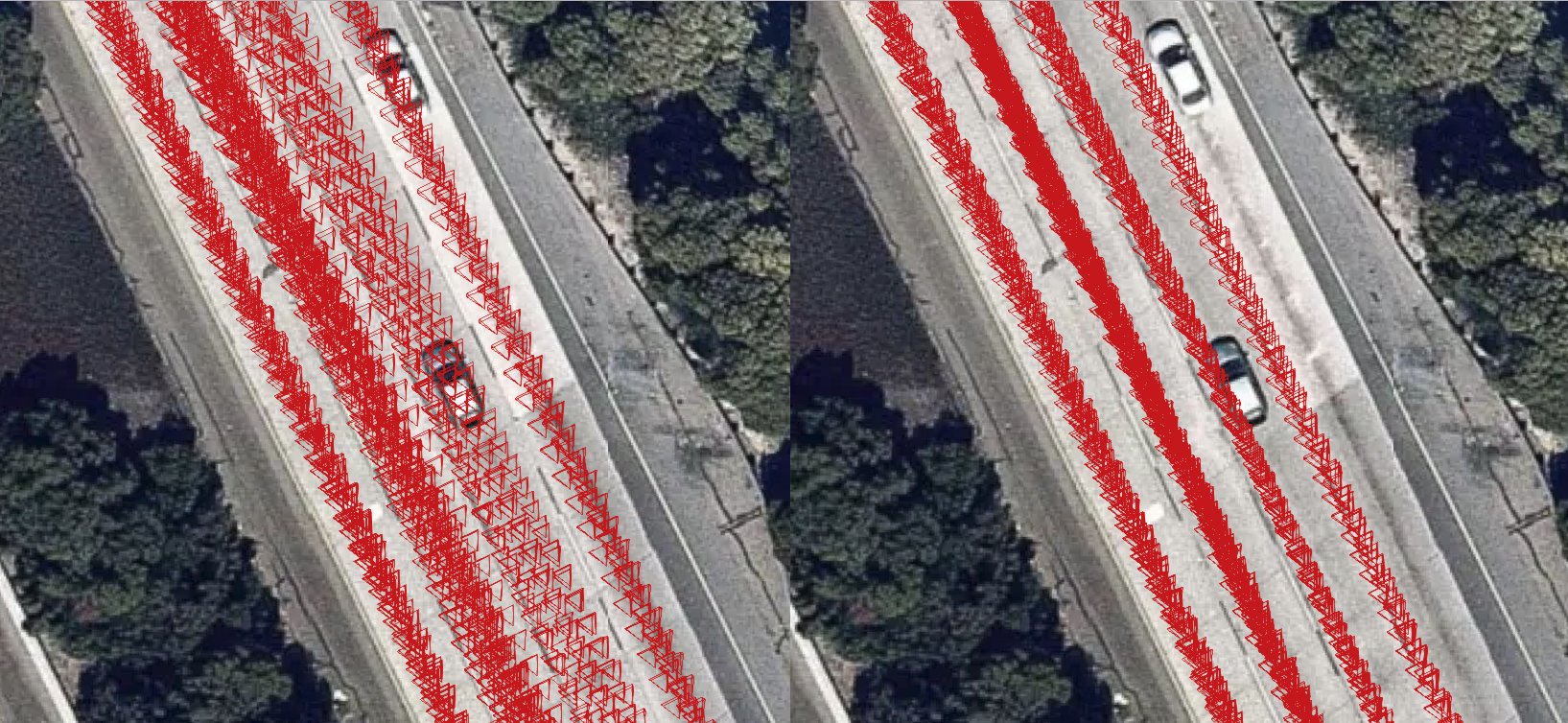}
  \caption{Viewing frusta from provided global poses (left) and global poses with map-based corrections (right).}
  \label{correction}
\end{figure*}

To estimate the Root Mean Squared Error (RMSE) of the vertical component of position, we used the same technique as in Section \ref{LAIKA}. By using the observed DOP\cite{langley1999dilution} of each fix we can get a reliable estimate of horizontal errors too. To estimate the accuracy of the provided orientation, we took the Jacobian, $J_{\Delta\theta_i} = \partial R_{i} / \partial \Delta\theta_i$, of the re-projection error $(R)$ for the $i_{th}$ observed ORB feature, with respect to orientation errors, $\Delta\theta$. We can then create linear equations to estimate the orientation error by using the Jacobian to linearize around $\Delta\theta_{i}=0$. The high level equations used to calculate the RMSE of the orientation, $\hat{\theta}$, are shown below (\ref{angleerros}).
\begin{equation}
\begin{array}{cc}
    \hat{\Delta\theta_{i}} = J_{\Delta\theta_{i}}^{-1} R_i \\
    RMSE(\hat{\theta}) = \sqrt{E\Big[\hat{\Delta\theta_{i}}^2\Big]}
\end{array}
\label{angleerros}
\end{equation}
Since most of the measured reprojection error, $R$, is due to noise in the ORB feature detection, it is fair to assume that (\ref{angleerros}) is an upper bound of the true orientation errors in our estimates. In Table \ref{pose_acc} we show both estimated position and orientation errors.

Some applications require even more accurate poses than provided above. One can use vision to fine tune the pose estimates with a simple Expectation-Maximization algorithm: first average the ECEF position of the matching ORB features across image/pose pairs from different drives, this reduces the error in ORB feature localization. After that, we infer the corrected poses by relocalizing the frames against the averaged ORB features. An example of a single iteration of this type of correction is show in Figure \ref{correction}.

\begin{table}[!htb]
    \caption{Estimated RMSE of provided global poses}
    \centering
    \begin{tabular}{|p{.8cm}|p{.8cm}|p{.8cm}|p{.8cm}||p{.8cm}|p{.8cm}|p{.8cm}|}
     \multicolumn{1}{c}{} & \multicolumn{3}{c}{Position RMSE} & \multicolumn{3}{c}{Orientation RMSE} \\
     \hline
     & North & East & Down & Roll & Pitch & Yaw\\
     \hline
     RAV4 & 0.6m & 0.6m & 0.9m &  $0.20^\circ$ & $0.20^\circ$ & $0.25^\circ$\\
    \hline
    Civic & 1.3m & 1.3m & 2m &  $0.20^\circ$ & $0.20^\circ$ & $0.25^\circ$\\
    \hline
    \end{tabular}
    \label{pose_acc}
\end{table}

\section{CONCLUSION}
We proposed the \texttt{comma2k19}, a state-of-the-art dataset to develop and validate tightly coupled GNSS algorithms, fused pose estimators and mapping algorithms that are intended to work with commodity sensors. Using \texttt{comma2k19} we built and open sourced Laika, a raw GNSS processing library that reduced positioning errors by 40\% compared to the baseline algorithm shipped with the u-blox sensor used data collection. \texttt{comma2k19} also includes camera poses in a global reference frame of the over 2 million images provided. We believe the most interesting future research directions using \texttt{comma2k19} and Laika should be developing novel vision and sensor fusion based mapping algorithms for HD maps in highways with sparse features to track.  

\section*{ACKNOWLEDGEMENT}
We'd like to thank Eddie Samuels, Nicholas McCoy, George Hotz, Greg Hogan, Viviane Ford and Willem Melching for setting up the hardware and infrastructure that enabled this research.

\bibliographystyle{unsrt}
\bibliography{papers}

\begin{thebibliography}{10}

\bibitem{schall2009global}
Gerhard Schall, Daniel Wagner, Gerhard Reitmayr, Elise Taichmann, Manfred
  Wieser, Dieter Schmalstieg, and Bernhard Hofmann-Wellenhof.
\newblock Global pose estimation using multi-sensor fusion for outdoor
  augmented reality.
\newblock In {\em Mixed and Augmented Reality, 2009. ISMAR 2009. 8th IEEE
  International Symposium on}, pages 153--162. IEEE, 2009.

\bibitem{rehder2012global}
Joern Rehder, Kamal Gupta, Stephen Nuske, and Sanjiv Singh.
\newblock Global pose estimation with limited gps and long range visual
  odometry.
\newblock In {\em 2012 IEEE international conference on robotics and
  automation}, pages 627--633. IEEE, 2012.

\bibitem{caron2006gps}
Francois Caron, Emmanuel Duflos, Denis Pomorski, and Philippe Vanheeghe.
\newblock Gps/imu data fusion using multisensor kalman filtering: introduction
  of contextual aspects.
\newblock {\em Information fusion}, 7(2):221--230, 2006.

\bibitem{nemra2010robust}
Abdelkrim Nemra and Nabil Aouf.
\newblock Robust ins/gps sensor fusion for uav localization using sdre
  nonlinear filtering.
\newblock {\em IEEE Sensors Journal}, 10(4):789--798, 2010.

\bibitem{angrisano2010gnss}
Antonio Angrisano.
\newblock Gnss/ins integration methods.
\newblock {\em Dottorato di ricerca (PhD) in Scienze Geodetiche e Topografiche
  Thesis, Universita’degli Studi di Napoli PARTHENOPE, Naple}, 21, 2010.

\bibitem{falco2017loose}
Gianluca Falco, Marco Pini, and Gianluca Marucco.
\newblock Loose and tight gnss/ins integrations: Comparison of performance
  assessed in real urban scenarios.
\newblock {\em Sensors}, 17(2):255, 2017.

\bibitem{schreiber2016vehicle}
Markus Schreiber, Hendrik K{\"o}nigshof, Andr{\'e}-Marcel Hellmund, and
  Christoph Stiller.
\newblock Vehicle localization with tightly coupled gnss and visual odometry.
\newblock In {\em Intelligent Vehicles Symposium (IV), 2016 IEEE}, pages
  858--863. IEEE, 2016.

\bibitem{geiger2013vision}
Andreas Geiger, Philip Lenz, Christoph Stiller, and Raquel Urtasun.
\newblock Vision meets robotics: The kitti dataset.
\newblock {\em The International Journal of Robotics Research},
  32(11):1231--1237, 2013.

\bibitem{cordts2016cityscapes}
Marius Cordts, Mohamed Omran, Sebastian Ramos, Timo Rehfeld, Markus Enzweiler,
  Rodrigo Benenson, Uwe Franke, Stefan Roth, and Bernt Schiele.
\newblock The cityscapes dataset for semantic urban scene understanding.
\newblock In {\em Proceedings of the IEEE conference on computer vision and
  pattern recognition}, pages 3213--3223, 2016.

\bibitem{maddern20171}
Will Maddern, Geoffrey Pascoe, Chris Linegar, and Paul Newman.
\newblock 1 year, 1000 km: The oxford robotcar dataset.
\newblock {\em The International Journal of Robotics Research}, 36(1):3--15,
  2017.

\bibitem{huang2018apolloscape}
Xinyu Huang, Xinjing Cheng, Qichuan Geng, Binbin Cao, Dingfu Zhou, Peng Wang,
  Yuanqing Lin, and Ruigang Yang.
\newblock The apolloscape dataset for autonomous driving.
\newblock {\em arXiv preprint arXiv:1803.06184}, 2018.

\bibitem{yu2018bdd100k}
Fisher Yu, Wenqi Xian, Yingying Chen, Fangchen Liu, Mike Liao, Vashisht
  Madhavan, and Trevor Darrell.
\newblock Bdd100k: A diverse driving video database with scalable annotation
  tooling.
\newblock {\em arXiv preprint arXiv:1805.04687}, 2018.

\bibitem{santana2016learning}
Eder Santana and George Hotz.
\newblock Learning a driving simulator.
\newblock {\em arXiv preprint arXiv:1608.01230}, 2016.

\bibitem{goodfeatures}
Jianbo Shi and Tomasi.
\newblock Good features to track.
\newblock In {\em 1994 Proceedings of IEEE Conference on Computer Vision and
  Pattern Recognition}, pages 593--600, June 1994.

\bibitem{takasu2011rtklib}
T~Takasu.
\newblock Rtklib: An open source program package for gnss positioning, 2011.

\bibitem{harris2007gpstk}
R~Benjamin Harris and Richard~G Mach.
\newblock The gpstk: an open source gps toolkit.
\newblock {\em GPS Solutions}, 11(2):145--150, 2007.

\bibitem{rublee2011orb}
Ethan Rublee, Vincent Rabaud, Kurt Konolige, and Gary Bradski.
\newblock Orb: An efficient alternative to sift or surf.
\newblock In {\em Computer Vision (ICCV), 2011 IEEE international conference
  on}, pages 2564--2571. IEEE, 2011.

\bibitem{mourikis2007multi}
Anastasios~I Mourikis and Stergios~I Roumeliotis.
\newblock A multi-state constraint kalman filter for vision-aided inertial
  navigation.
\newblock In {\em Robotics and automation, 2007 IEEE international conference
  on}, pages 3565--3572. IEEE, 2007.

\bibitem{li2013high}
Mingyang Li and Anastasios~I Mourikis.
\newblock High-precision, consistent ekf-based visual-inertial odometry.
\newblock {\em The International Journal of Robotics Research}, 32(6):690--711,
  2013.

\bibitem{zhou1999sensor}
Yifeng Zhou, Henry Leung, and Martin Blanchette.
\newblock Sensor alignment with earth-centered earth-fixed (ecef) coordinate
  system.
\newblock {\em IEEE Transactions on Aerospace and Electronic systems},
  35(2):410--418, 1999.

\bibitem{cai2011unmanned}
Guowei Cai, Ben~M Chen, and Tong~Heng Lee.
\newblock {\em Unmanned rotorcraft systems}.
\newblock Springer Science \& Business Media, 2011.

\bibitem{langley1999dilution}
Richard~B Langley et~al.
\newblock Dilution of precision.
\newblock {\em GPS world}, 10(5):52--59, 1999.

\end{thebibliography}
\end{document}